\documentclass[letterpaper, 10 pt, conference]{support/ieeeconf}
\usepackage{times}
\usepackage[pdftex]{graphicx}
\usepackage{subfigure}
\usepackage{amsmath,amssymb,amsopn,amstext,amsfonts}
\usepackage{cancel}
\usepackage[space]{cite}
\usepackage{pdfsync}
\usepackage{balance}
\usepackage{color}
\usepackage{mathtools}
\usepackage{algpseudocode}
\usepackage{algorithm} 

\algnewcommand\algorithmicforeach{\textbf{for each}}
\algdef{S}[FOR]{ForEach}[1]{\algorithmicforeach\ #1\ \algorithmicdo}

\usepackage{bm}

\usepackage{diagbox}
\usepackage{epstopdf}
\usepackage{pifont}
\usepackage{fixltx2e}
\usepackage{amsmath}
\usepackage{multirow}
\usepackage{url}
\usepackage{verbatim}
\usepackage[linkcolor=black,citecolor=black,urlcolor=black,colorlinks=true]{hyperref}
\usepackage{subfigure}

\graphicspath{{figures/}}
\DeclareGraphicsExtensions{.png,.jpg,.eps,.pdf}
\IEEEoverridecommandlockouts
\newcommand{\norm}[1]{\left\lVert#1\right\rVert}
\title{\LARGE \bf
Flying through a narrow gap using neural network: an end-to-end planning and control approach \& Symposia*
}

\author{Albert Author$^{1}$ and Bernard D. Researcher$^{2}$
\thanks{*This work was not supported by any organization}
\thanks{$^{1}$Albert Author is with Faculty of Electrical Engineering, Mathematics and Computer Science,
        University of Twente, 7500 AE Enschede, The Netherlands
        {\tt\small albert.author@papercept.net}}%
\thanks{$^{2}$Bernard D. Researcheris with the Department of Electrical Engineering, Wright State University,
        Dayton, OH 45435, USA
        {\tt\small b.d.researcher@ieee.org}}%
}

\begin{document}
	
	\title{Flying through a narrow gap using neural network: an end-to-end planning and control approach}
	
	\author{Jiarong Lin, Luqi Wang, Fei Gao, Shaojie Shen and Fu Zhang
	\thanks{
	J. Lin and F. Zhang are with the Department of Mechanical Engineering, Hong Kong University, Hong Kong SAR., China. {\tt\small $\{$jiarong.lin,  fuzhang$\}$@hku.hk}
	L. Wang, F. Gao,  and S. Shen are with the Department of Electronic and Computer Engineering, Hong Kong University of Science and Technology, Hong Kong SAR., China. {\tt\small  lwangax@connect.ust.hk, $\{$fgaoaa, eeshaojie$\}$@ust.hk.}
	}	
}%


	
	

	\maketitle
	
	\begin{abstract}
         In this paper, we investigate the problem of enabling a drone to fly through a tilted narrow gap, without a traditional planning and control pipeline. To this end, we propose an end-to-end policy network, which imitates from the traditional pipeline and is fine-tuned using reinforcement learning. Unlike previous works which plan dynamical feasible trajectories using motion primitives and track the generated trajectory by a geometric controller, our proposed method is an end-to-end approach which takes the flight scenario as input and directly outputs thrust-attitude control commands for the quadrotor. 

		Key contributions of our paper are: 1) presenting an imitate-reinforce training framework. 2) flying through a narrow gap using an end-to-end policy network, showing that learning based method can also address the highly dynamic control problem as the traditional pipeline does (see attached video\footnote{\url{https://www.youtube.com/watch?v=jU1qRcLdjx0}}). 3) propose a robust imitation of an optimal trajectory generator using multilayer perceptrons. 4) show how reinforcement learning can improve the performance of imitation learning, and the potential to achieve higher performance over the model-based method.
		
	\end{abstract}
	\IEEEpeerreviewmaketitle
	
 	\section{Introduction}
		In the field of mobile robots, the paradigm of state-of-the-art work \cite{lin2018autonomous, gaoflying} addressing the autonomous navigation and control problem is perception-planning-control. In this paradigm, we first estimate the robot state and build a map of its surrounding environment by means of Simultaneous Localization and Mapping (SLAM). Within this map, a smooth, optimal trajectory is usually planned and executed via a low-level tracking controller. This approach is easy to analyze by well separating the design, analysis, and optimization of each module within the pipeline, and has proven very successful in many robotic applications, especially in low-speed, static environments. However, for aggressive robot maneuvers in cluttered, dynamic environments, such as drones racing in bush or indoor scenario, this approach becomes quite challenging because SLAM and trajectory optimization is memory and computationally expensive and degrade in performance for aggressive, dynamic maneuvers in non-static environments. 
 							
        More recently, end-to-end approaches \cite{hwangbo2019learning} have been proposed to achieve more aggressive robots maneuvers in cluttered dynamic environments. The basic idea is to train a control policy that directly maps sensory inputs to control outputs. Due to the shorter pipeline and its neural network structured policy controller, an end-to-end approach has the potential to achieve less computation time by utilizing the parallel computation of current GPUs.  It could also mitigate the accumulation modeling error contributed by each module within the conventional pipeline, by optimizing the end-to-end policy network globally\cite{hwangbo2019learning}. 
        
       	Despite these benefits, the end-to-end approach suffers from two major drawbacks: (1) training of the policy network typically requires a reinforcement learning framework, which improves the network parameters using data collected in trial tests. As the policy network becomes more complicated, the needed training data (thus trial tests) grows exponentially. (2) the trained policy network has no mathematical proof on its stability nor robustness. 
          
    	In this work, we investigate the stability and robustness of the end-to-end control approach in aggressive drone flights. We consider the drone flying through a narrow gap at a maximum speed up to $ 3 m/s $, and orientation angle up to $ 60^\circ $, such scenario poses an extremely high requirement on both the precision and robustness of the control policy. We start with replacing the traditional model-based motion planner and tracking controller with a neural network based policy controller. This policy network takes the mapping results as input and directly computes the control actions. Experiment results that such a neural-network-based control policy is indeed able to achieve comparable accuracy and stability with conventional motion planner and tracking controller. What’s more, our network fine-tuned by reinforcement learning is outperforms traditional model-based method in some properties, indicating the potential of our imitate-reinforce framework can achieve higher performance over the model-based approach. 
    	To share our finding with robotics community, we will publicly release our codes, trained network, and simulator\footnote{\url{https://github.com/hku-mars/crossgap_il_rl}}.

%
	
\begin{figure}[t]
		\vspace{-0.2cm}
	{\includegraphics[width=0.95\columnwidth]{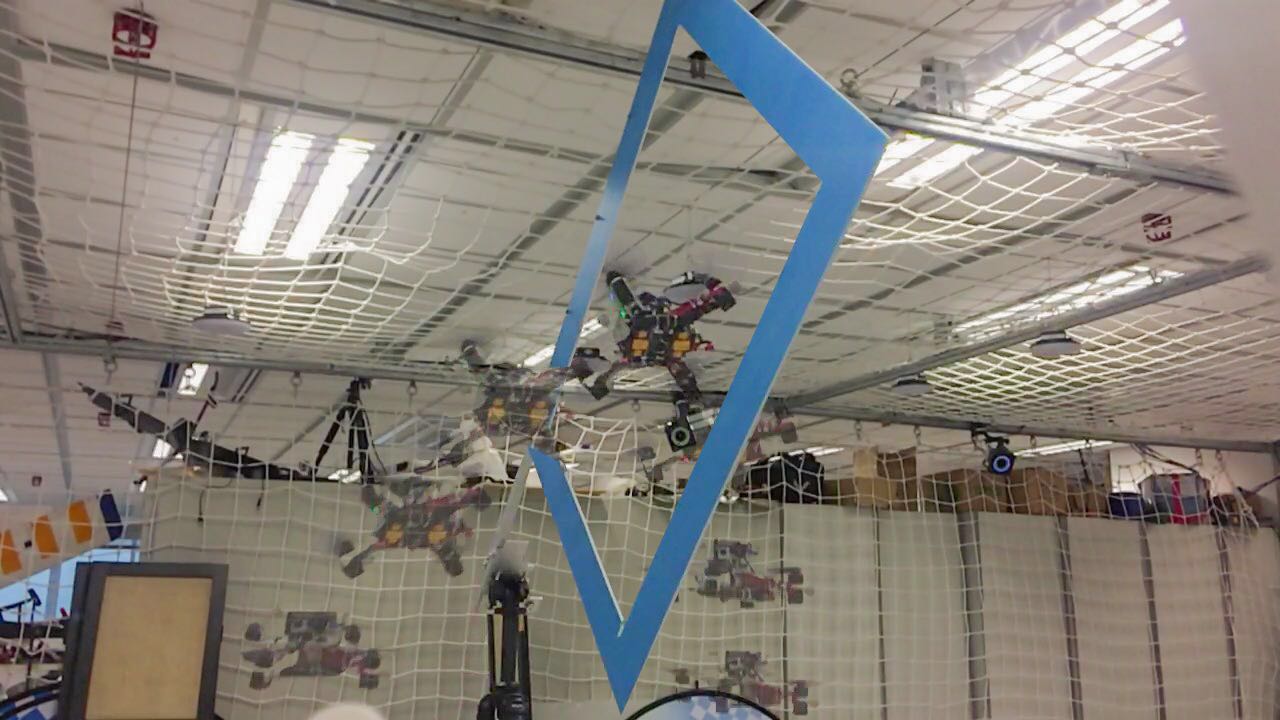}}
	\caption{Our quadrotor flying through a narrow gap
		\label{fig:view}
	}
	\vspace{-1.5cm}
\end{figure}

\begin{figure*}[htp]
	\centering
	{\includegraphics[width=2.0\columnwidth]{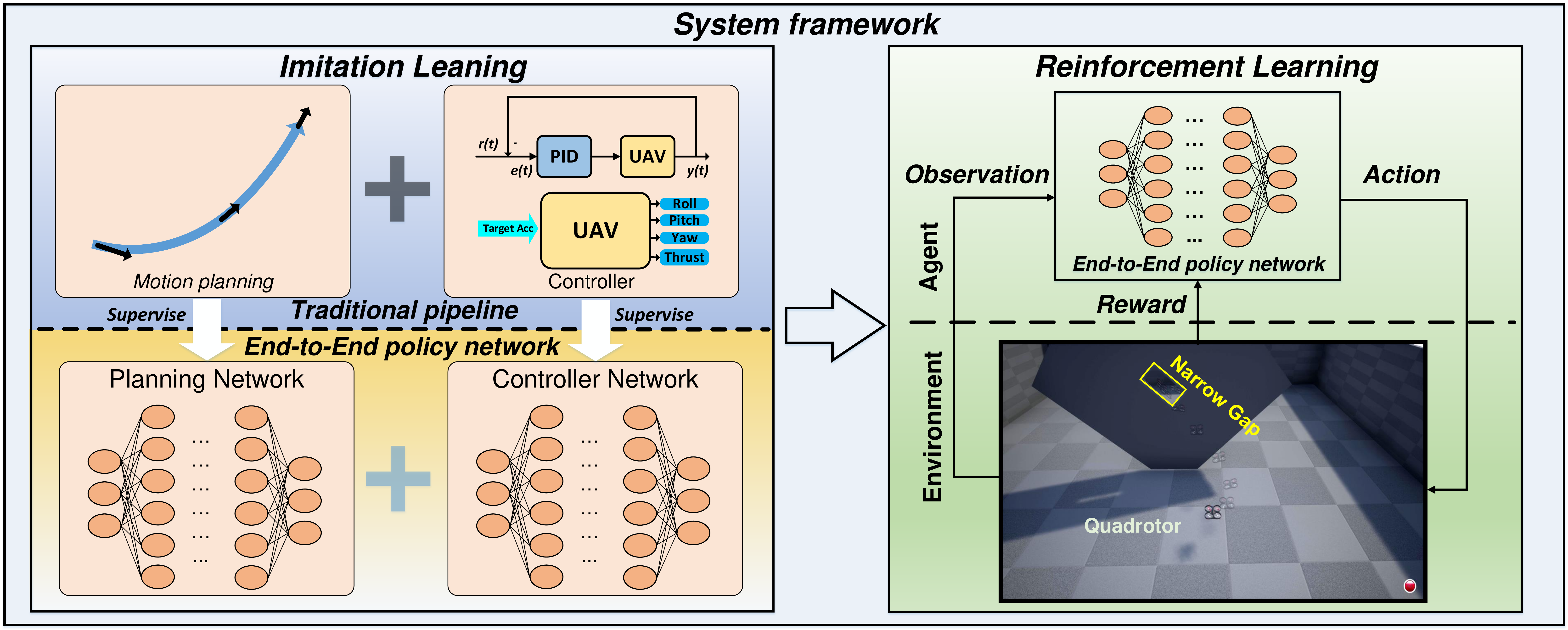}}
	\caption{
		The framework of our work can be divided into two phases, the imitation and reinforcement learning. In the first phase, we train our end-to-end policy network by imitating from a tradition pipeline. In the second phase, we fine-tune our policy network using reinforcement learning to improve the network performance.
		\label{fig:framework}
	}
		\vspace{-0.5cm}		
\end{figure*}
	\section{Related work}
	
        
     With the development of deep learning technology, the learning-based methods are playing a more and more significant role in the field of autonomous navigation for mobile robots. For example, Giusti, \textit{et al}. in \cite{giusti2016machine} propose a learning-based visual perception which enabled the quadrotor flying on forest trails automatically. In \cite{jung2018perception}, authors facilitate the drone safety fly in dynamic environments with perception provided from deep-neural networks. Kaufmann, \textit{et al}. \cite{kaufmann2018deep, kaufmann2018beauty} show that combining learning-based method with traditional methods can successfully fly with high agility in Drone Racing. Reinforcement learning is applied in addressing the challenging problem of helicopter's aerobatic flights \cite{abbeel2007application}. These works suggest that learning-based methods are effective ways to deal with the problems in the UAV (unmanned aerial vehicle) flights. 
     
	Aggressive flight through a narrow gap is one of the most challenging problems in autonomous quadrotors control. To minimize the risk of collision, it requires the quadrotor to pass through the center with its attitude aligned with the orientation of the gap.  In \cite{mellinger2012trajectory}, authors achieve the goal by tracking the sequence of trajectories designed offline. Mellinger, \textit{et al.} in \cite{loianno2017estimation} consider the autnomous navigation using state estimation form a monocular camera and an IMU. Falanga, \textit{et al}. in \cite{falanga2017aggressive} further accomplish the goal without any prior knowledge of the pose of the gap, using only onboard sensing and computing. Takes these work as a baseline, we investigate the feasibility and performance of end-to-end approach.

	\section{ Imitate-reinforce training framework }
        In our work, addressing the problem of flying through a narrow gap using an end-to-end neural network, we first learn the function of the traditional pipeline by using two neural networks imitating the traditional motion planning and controller. After imitation learning, we fine-tune the neural network using reinforcement learning to improve its performance. The whole framework of our system is shown in  Fig.~\ref{fig:framework}

	\section{Imitation of motion planning} \label{Sect_imitate_planning}
	In this section, we will introduce how we use multilayer perceptrons (MLP) to imitate a motion primitive generator, including the design of neural-network, learning of cost function and data normalization.
	\subsection{Problem statement}
	Imitating a motion primitive generator \cite{mueller2015computationally} can be viewed as using multilayer Perceptrons (MLP) to regress it. According to the universal approximation theorem \cite{hornik1989multilayer, cybenko1989approximation}, we could use a large MLP to approximate a very complicated function, where the approximation accuracy will depend on the size of the MLP \cite{goodfellow2016deep}.
	
	For a quadrotors traveling from a starting state $\mathbf{S}_{s}$ (including position $ \mathbf{p}_s$, velocity $ \mathbf{v}_s $ and $ \mathbf{a}_s $) to an ending state $\mathbf{S}_{e}$ ($ \mathbf{p}_e, \mathbf{v}_e$ and $\mathbf{a}_e $) with time duration $T$. The motion primitive generator in \cite{mueller2015computationally} generates an average jerk optimal trajectory by utilizing the Pontryagin's maximum principle. After generating the trajectory, we obtain the desired position $ \mathbf{p}(t) $, velocity $ \mathbf{v}(t) $ and acceleration $\mathbf{a}(t)$ for controlling the quadrotor, where $ t \leq T$.

	\subsection{Network structure}
\begin{figure}[h]
		\centering
		{\includegraphics[width=1.0\columnwidth]{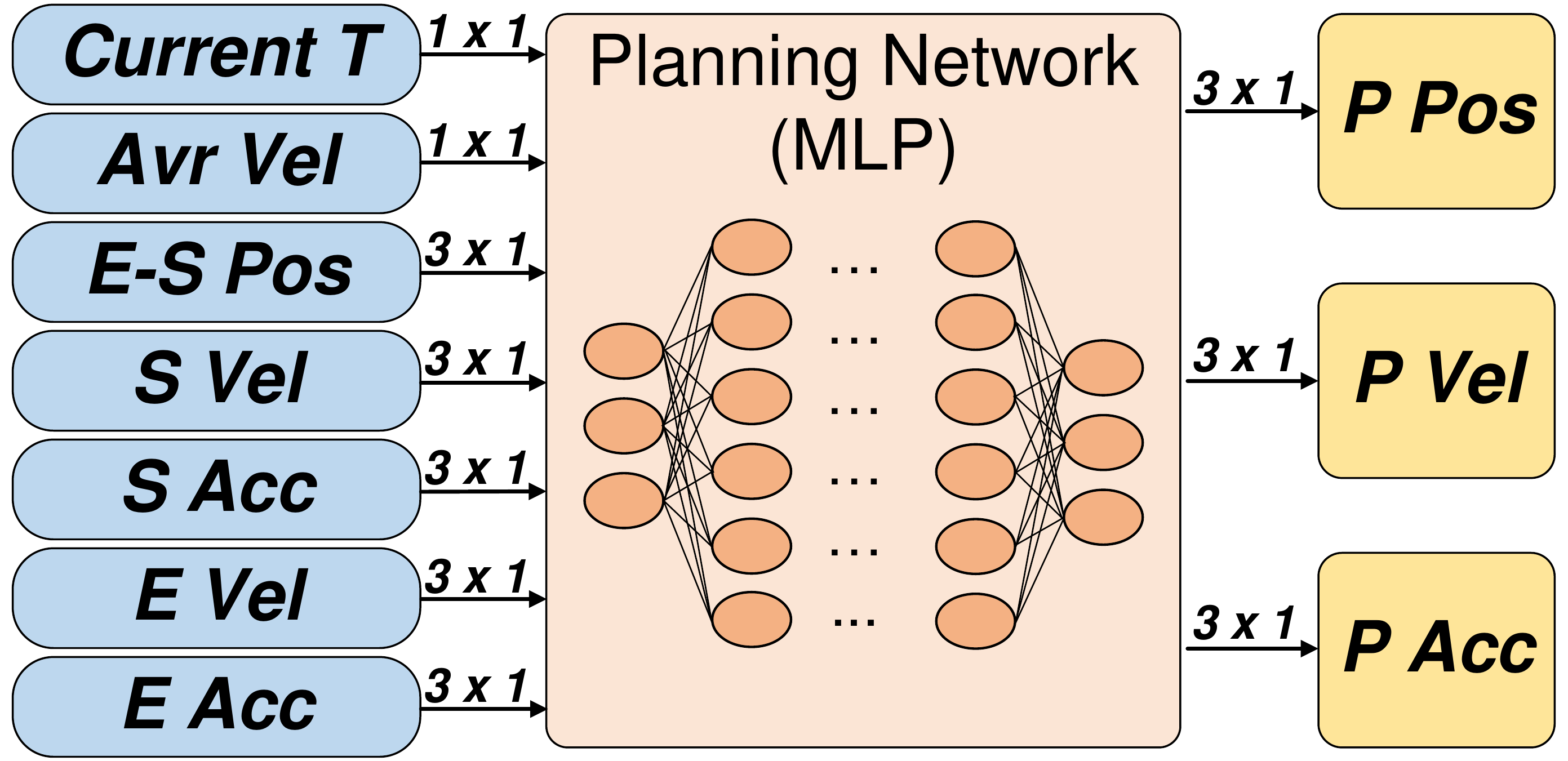}}
		\caption{
			Input and output of our planning network. The input is a $17 \times 1$ vector including Current Time  $t$ (\textit{Current T}, ${1\times 1}$), Average Velocity $\bar{v}$(\textit{Avr Vel}, $ 1 \times 1 $), Start to the End vector relative Position  $\Delta\mathbf{p}_{s\rightarrow e} $(\textit{E-S Pos}, $ 3\times 1 $), Starting Velocity  $\mathbf{v}_s$ (\textit{S Vel}, $ 3\times 1 $), Starting Acceleration  $ \mathbf{a}_s $ (\textit{S Acc}, $ 3\times 1 $), Ending Velocity  $ \mathbf{v}_e $ (\textit{E Vel}, $ 3\times 1 $) and Ending Acceleration $ \mathbf{a}_e $ (\textit{E Acc}, $ 3\times 1 $). The output is a $9 \times 1$ vector, including the prediction of relative position $ \Delta\mathbf{p}_p$ (\textit{P Pos}, $3\times 1$), velocity $ \mathbf{v}_p$ (\textit{P Vel}, $3\times 1$) and acceleration prediction  $ \mathbf{a}_p$ (\textit{P Acc}, $3\times 1$)
			\label{fig:planning_network_in_out}}
			\vspace{-0.8cm}
	\end{figure}

	In this paper, the designed MLP framework is shown in Fig.~\ref{fig:planning_network_in_out}. The input of the network is a $ 17 \times 1 $ vector, and the output of the network is a $ 9 \times 1 $ vectors. As for the planning network, it has 10 fully-connected layers and each layer has $ 100 $ latent units (shown in Fig.~ \ref{fig:planning_network}).
	
	\subsection{Network Training}
	\subsubsection{Data collection}
	We generate 20 thousand trajectories by using a random set of starting and ending states as training samples. Each trajectory is discreted to 1000 points uniformly distributed between $0$ and $T$, where $T = || \Delta\mathbf{p}_{e\rightarrow s}||/\bar{v}$ also called the traveling time. The start to the end relative position $  \Delta\mathbf{p}_{s\rightarrow e} $ in each single axis lies in $ -30\sim 30 m $, velocity ($\mathbf{v}_s, \mathbf{v}_e$) and acceleration ($\mathbf{a}_{s}, \mathbf{a}_e$) are in $  -10\sim 10 m/s $ and $ -10\sim 10 m/s^2 $, respectively. The average velocity $\bar{v} $ ranges in $1\sim7 m/s$. On the consideration of the training stability, we manually remove those trajectories with too large outputs.
	
	\subsubsection{Loss function}
	We train our neural network with a weighted MSE loss on position, velocity, and acceleration. The loss-function is:
	$$
	\begin{aligned}
	Loss = & w_{p}\cdot{|| \Delta\mathbf{p}_{l} - \Delta\mathbf{p}_{p} ||}^2 + w_{v}\cdot {|| \mathbf{v}_{l} - \mathbf{v}_{p} ||}^2 \\
	+ &w_{a}\cdot {|| \mathbf{a}_{l} - \mathbf{a}_{p} ||}^2 + g \\
	\end{aligned} 
	$$
	where $\Delta\mathbf{p}_{l}, \mathbf{v}_{l}, \mathbf{a}_{l}  $ are the relative position (relate to starting position), velocity, acceleration of labeling data generated from a conventional motion planner. $g$ is the weight-decay factor which can improve the generalization capability of our network.
	
 	To enhance the flying safety, we consider that the position error is the most important item and therefore should be assigned with the highest weight. Then, the velocity should set as the second place, and the last is the acceleration. In  our work, the weigh $w_{p} $, $w_{p} $ and $ w_{a}$ are set as $4$, $2$ and $1$, respectively.
	
	\begin{figure}[t]
		\subfigure[\label{fig:planning_network} Plannning network ]
		{\centering\includegraphics[width=0.38\columnwidth]{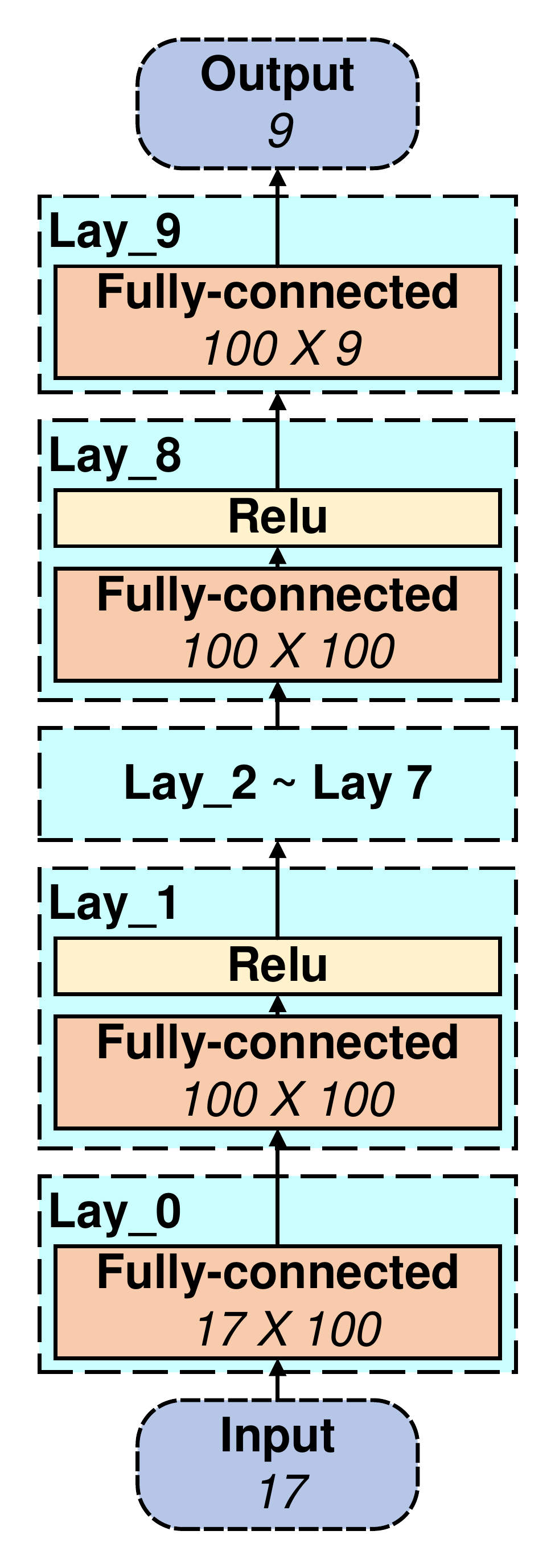}}			\subfigure[\label{fig:planning_network_with_scale} Plannning network with data normalization ]
		{\centering\includegraphics[width=0.60\columnwidth]{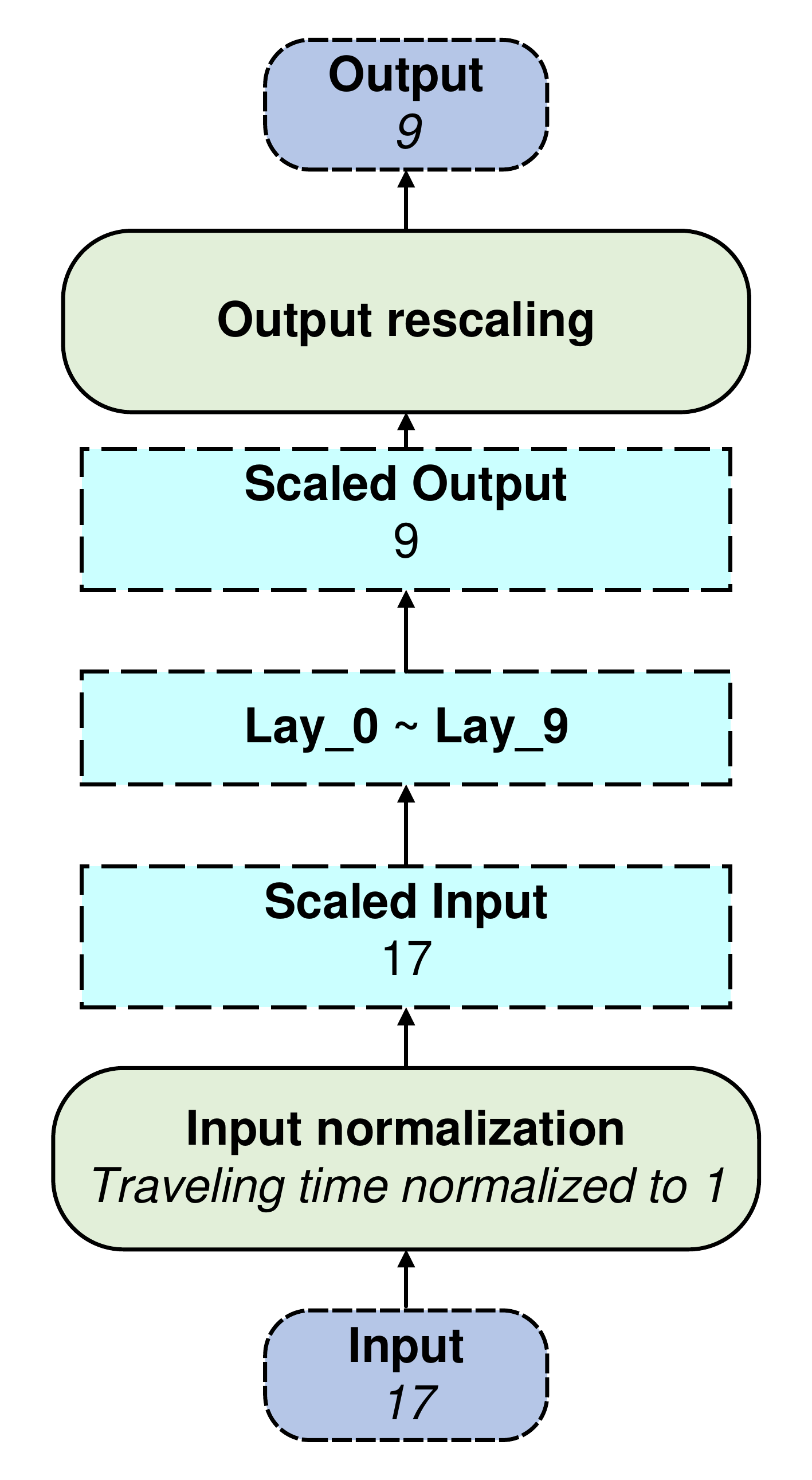}}
		\vspace{-0.8cm}
	\end{figure}
	
	\subsection{Data normalization}
	    In \cite{sola1997importance, ioffe2015batch, singh2015investigations}, authors show that data normalization plays an important role in achieving a satisfactory result in the training process.  In our work, we normalize our traveling time $T$ to 1 to accelerate the training process and improve the precision of imitation learning. 
	
	We scale the input and re-scale the output data of the MLP-network (shown in Fig.~\ref{fig:planning_network_with_scale}). The scale factor $s$ is equal to the traveling time $T$.
	$$ s = T $$ 
	
	The scaled time $ t' = t/s $ and relative position (relative to $ \mathbf{p}_s $)  $\Delta\mathbf{p}'(t') = s\cdot \Delta\mathbf{p}(t) $, we have
	$$
	\begin{aligned}
	v'(t') &= \dfrac{d}{t'}\mathbf{p}'(t') = s^2 v(t) \\
	a'(t') &= \dfrac{d}{t'}\mathbf{v}'(t') = s^3 a(t)
	\end{aligned}	
	$$	
	
	By this, the scaled inputs vector is given as below
	$$
	\begin{aligned}
	t' = t/s, ~\bar{v}' = s^2 \cdot \bar{v},&~ \Delta\mathbf{p}'_{e-s} = s \cdot \Delta\mathbf{p}_{s\rightarrow e}, \\
	\mathbf{v}'_{s} = s^2\cdot \mathbf{v}_{s},&~~ \mathbf{v}'_{e} = s^2\cdot \mathbf{v}_{e} \\
	\mathbf{a}'_{s} = s^3\cdot \mathbf{a}_{s},&~~ \mathbf{a}'_{e} = s^3\cdot \mathbf{a}_{e} \\
	\end{aligned}
	$$
	
	Correspondently, the re-scaled outputs is 
	$$
	\begin{aligned}
	\Delta\mathbf{p}_{p} = \Delta\mathbf{p}'_{p}/s, ~\mathbf{v}_{p} = \mathbf{v}'_{p}/s^2,~\mathbf{a}_{p} = \mathbf{a}'_{p}/s^3 
	\end{aligned}
	$$
	\subsection{Data augmentation}\label{section_data_argu}
	
	For a pair of raw training data, including input data: $\{ t, \Delta\mathbf{p}_{s\rightarrow e}, \bar{v}, \mathbf{v}_s, \mathbf{a}_s, \mathbf{v}_e, \mathbf{a}_e \}$ and output data $\{\Delta \mathbf{p}_l(t), \mathbf{v}_l(t), \mathbf{a}_l(t) \}$, we augment it in two ways enabled by the linearity property of the system.
	\begin{itemize}	
	\item Sign flipping: We augment the data by flipping the sign of the data, the inputs of the augmentation data become:
	$$
	\begin{aligned}
	t' &= t,~ \Delta\mathbf{p}'_{e-s} = -\Delta\mathbf{p}_{s\rightarrow e},~ \bar{v}' = \bar{v} \\
	\mathbf{v}'_s = -\mathbf{v}_s,&~\mathbf{a}'_s = -\mathbf{a}_s, ~\mathbf{v}'_e = -\mathbf{v}_e, ~\mathbf{a}'_e = -\mathbf{a}_e,
	\end{aligned}
	$$
	
	and the output of augmentation data is flipped in the same way.
	\item Scaling: We augment the data by multiplying a random scale $ s~(s\leq 5)$ on both of the input and output data. The inputs of the augmentation data become:
		$$
	\begin{aligned}
	t' &= t,~ \Delta\mathbf{p}'_{e-s} = -s\Delta\mathbf{p}_{e\rightarrow s},~ \bar{v}' = s\bar{v} \\
	\mathbf{v}'_s = -s\mathbf{v}_s,&~\mathbf{a}'_s = -s\mathbf{a}_s, ~\mathbf{v}'_e = -s\mathbf{v}_e, ~\mathbf{a}'_e = -s\mathbf{a}_e,
	\end{aligned}
	$$
	and all the output of argumentation data should multiply the same scale factors too.
	\end{itemize}
	
	\section{Imitation of controller }
	Similar to the previous section, we will show how we use MLP to imitate a traditional controller, including the design of network structure, learning of cost function and so on.
	\subsection{Traditional controller}
	The traditional geometry tracking controller on $ SE(3) $ we imitate is in \cite{mellinger2011minimum, lee2010geometric}. In world frame coordinate  $ \mathcal{W} $ (shown in Fig.~\ref{fig:env_setup}), the current position , velocity, acceleration, and attitude of drone are denoted as $^w\mathbf{p}_c, ^w\mathbf{v}_c,  ^w\mathbf{a}_c$ and $^w\mathbf{R}_c$, respectively.  Given the desired position $ ^w\mathbf{p}_d $, velocity $^w\mathbf{v}_d$ , acceleration $ ^w\mathbf{a}_d $ and desired yaw angle $\psi_d$ , the controller can computes the desired roll $\phi_d$, pitch $\theta_d$ angle and thrust $\mu_d$.
	
	In our situation, our desired yaw direction is set as the $ X $-axis of the world frame ($ \psi_d \equiv 0 $), and the desired rotation matrix $ \mathbf{R}_d $ of UAV in the in world frame coordinate $ \mathcal{W} $ is (rotate in $X-Y-Z$ order )
	\begin{align}
	\mathbf{R}_d(\phi, \theta)  &= 
	\begin{bmatrix}
	1 & 0 & 0 \\
	0 & \cos \phi   & \sin \phi  \\
	0 &-\sin \phi   & \cos \phi  \\
	\end{bmatrix}
	\begin{bmatrix}
	\cos \theta  & 0 & -\sin \theta   \\
	0 & 1 & 0\\
	\sin \theta   & 0 & \cos \theta \\
	\end{bmatrix} \mathbf{I}_{3\times 3}  \nonumber \\
	&=
	\begin{bmatrix}
	\cos \theta  & 0 & -\sin \theta   \\
	\sin \phi    \cos \theta  &    \cos \phi  & \cos \theta  \sin \phi   \\
	\cos \phi    \sin \theta  & -\sin \phi  & \cos \phi  \cos \theta \\
	\end{bmatrix} \label{EQ_Rw_phi_theta}
	\end{align}
	
	%
	%
	
	\subsection{Network Structure}
\begin{figure}[h]
		\centering
z		{\includegraphics[width=1.0\columnwidth]{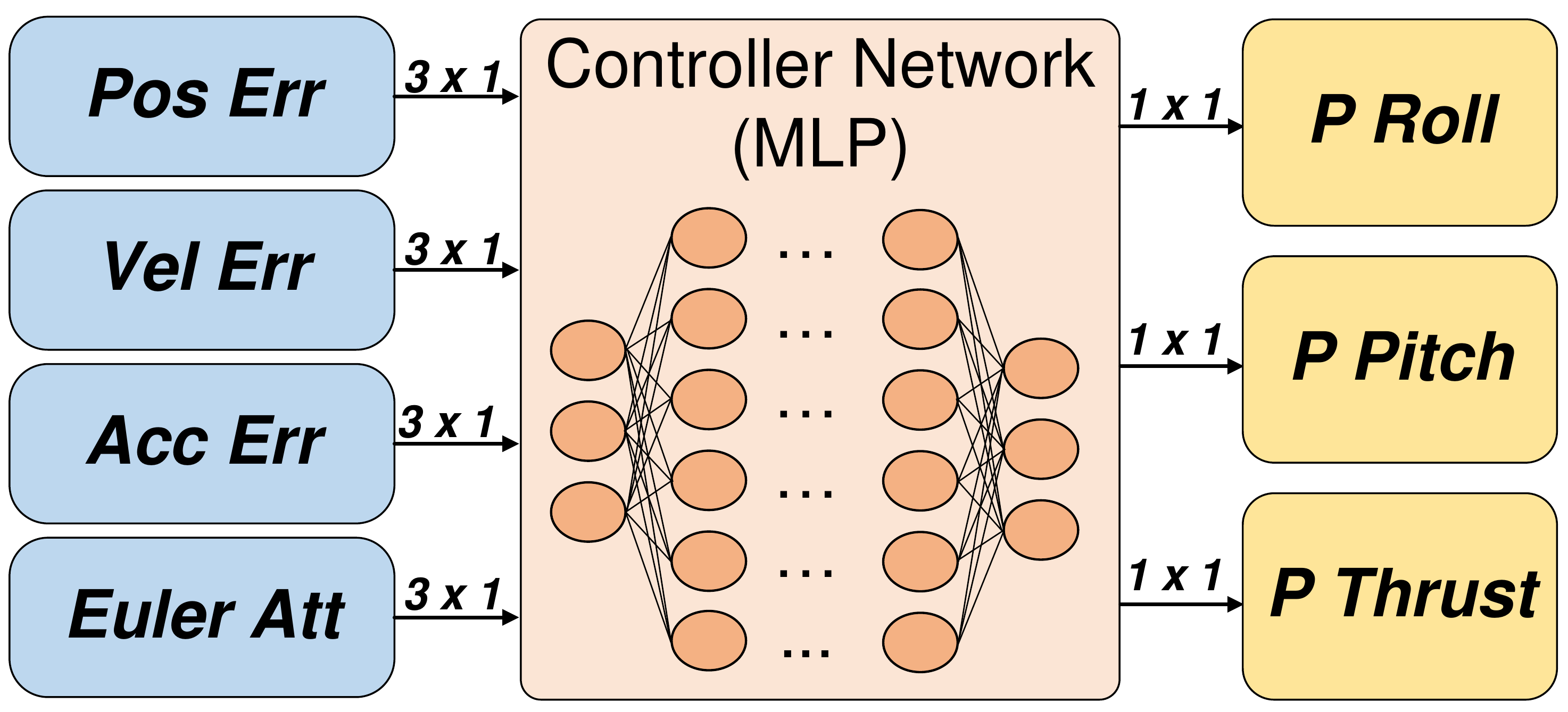}
			\caption{Input and output of our planning network. The input is a $12 \times 1$ including position error $ ^w\mathbf{e}_p $(\textit{Pos Err}, ${3\times 1}$), velocity error $ ^w\mathbf{e}_v $ (\textit{Vel Err}, $ 3 \times 1 $), acceleration error $ ^w\mathbf{e}_a $ (\textit{Acc Err}, $ 3\times 1 $), euler angle (roll $ \phi$, pitch $\theta $ and yaw $\psi$ angle) and attitude in Euler angles (\textit{Euler Att}, $ 3\times 1 $). The output of the network is a $ 3 \times 1 $ vector, including the predictions of roll  $\phi_p $ (\textit{P Roll}, $ 1\times 1 $),  pitch $\theta_p$ (\textit{P Pitch}, $ 1\times 1 $), and thrust $ \mu_p $ ( \textit{P Thrust}, $1 \times 1$).}
			\label{fig:controller_network_in_out}}
			\vspace{-0.4cm}
	\end{figure}

	The structure of the network is shown in Fig.~\ref{fig:controller_network_in_out}. The input of the network is a $ 12 \times 1 $ vector and the output of the network is a $ 3 \times 1 $ vector.

	In our work, the controller-network has the same number of latent layers of planning-network (shown in Fig.~\ref{fig:planning_network}). However, due to the lower dimensions of input and outputs, we reduce the number of latent units from $ 100 $ to $ 40 $, 
	
	
	\subsection{Network Training}
	\subsubsection{Data collection}
	We collect our training data by generating a large number of random input vectors and labeling their correspondent outputs using traditional cascaded PID controller. In our work, we generate two sets of training data, where each set of data contains $ 6\times 10^6 $ training samples. The difference between these two sets of data is their range of inputs. The first set of data contains a large range of inputs and is called Large-range dataset, the second set of data contains a short range of inputs vector and is therefore called Short-range dataset.
	\begin{itemize}
		\item Large-range dataset:  In this dataset, each axis of position error $ \mathbf{e}_p $ range in $-10 \sim 10 m $, Euler angle in $ -180 \sim 180 ^\circ$, velocity $ \mathbf{e}_v $ and acceleration $ \mathbf{e}_a $ in $ -5 \sim 5m/s$ and $ -10 \sim 10 m/s^2 $, respectively. Although our controller normally does not work under such kind of condition, we hope our MLP network can handle the large range of input error as well as the traditional method does, to increase its robustness to extreme cases.
		
		\item Short-range (working-range) dataset: In this dataset, each axis of position error $ \mathbf{e}_p $ lies in $ -0.2 \sim 0.2 m $, Euler angle in $ -30\sim 30^\circ $, velocity $ \mathbf{e}_v $ and acceleration $ \mathbf{e}_a $ in $ -0.3 \sim 0.3 m/s$ and $ -10 \sim 10 m/s^2 $, respectively. This range of input is the working situation of our controller, to guarantee the performance of the controller network, we add this dataset to the training data as well. 
	\end{itemize}
	\subsection{Loss function}
    We train our controller-network with a weighted MSE loss on thrust and Euler angle error.  The cost function is shown as follows.
	$$
	Loss  =  w_{thr}\cdot |\mu_l - \mu_p| + w_{eul}\cdot e_{l,p} + g
	$$
	where $g$ is the weight-decay factor, $ \mu_l $ is the output thrust of labeled data,  $ w_{thr} ,  w_{eul} $ are the weight factor of thrust and euler angle error $e_{l,p}$.
	
	The Euler angle error $e_{l,p}$ between labeling outputs $ \phi_l , \theta_l $ and predicting outputs $ \phi_p, \theta_p $ is: 
	$$
	\begin{aligned}
	e_{l,p} &= acos \left( \dfrac{ \text{tr} [ \mathbf{R}_d(\phi_l,\theta_l) \mathbf{R}_d^T(\phi_p,\theta_p) ] -1}{2} \right) \\
	\end{aligned}
	$$
	where, $ \mathbf{R}_d(\phi_l,\theta_l) $ and $ \mathbf{R}_d^T(\phi_p,\theta_p) $ are computed form Eq.~(\ref{EQ_Rw_phi_theta})

	In our work, the weight factor $ w_{thr} ,  w_{eul} $ are set to $ 1.0 $ and $ 57.3 $, respectively.
		
	\section{End-to-end planning and control}
\begin{figure}[h]
		\centering
		{\includegraphics[width=1.0\columnwidth]{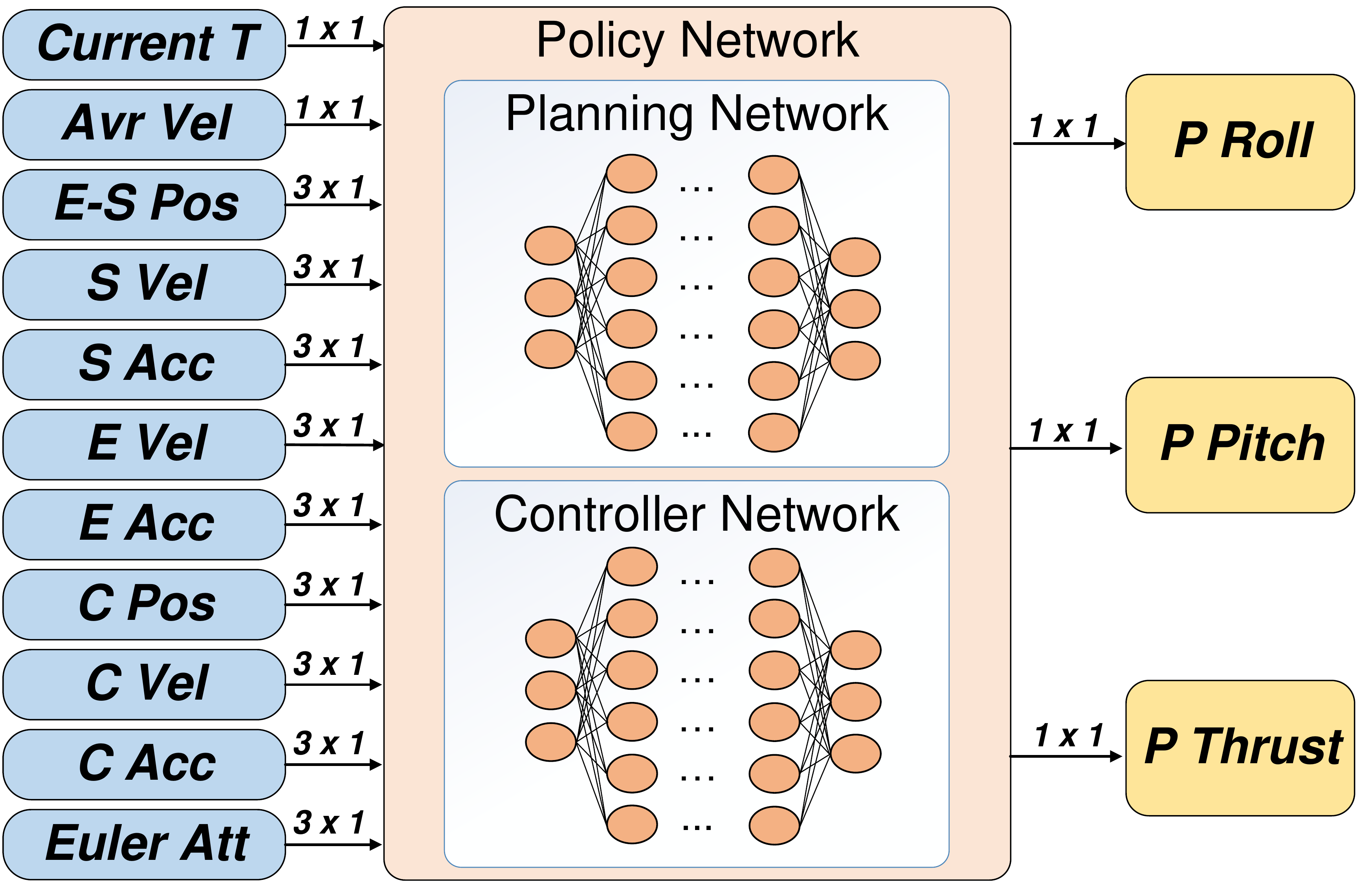}
			\caption{Input and output of end-to-end policy network. Where \textit{C Pos}, \textit{C Vel}, \textit{C Acc} is the current relative position $ \Delta \mathbf{p}_c $, $ \Delta \mathbf{v}_c $ and $ \mathbf{a}_c $, respectively.} 
			\label{fig:policy_network_in_out}}
			\vspace{-0.8cm}
	\end{figure}

	After imitating the traditional motion planning and controller individually, we can merge these two networks (shown in Fig.~\ref{fig:policy_network_in_out}), called the ``policy network". Given the observation of gap pose and the current state of the quadrotor, the policy network outputs the control command directly as traditional pipeline does.
	
	The input of the policy network is a $ 29 \times 1 $ vector including $ 17 \times 1 $ input for planning network and $ 12 \times 1 $ of current state. The input of the controller network is the output of planning network (including $ \Delta \mathbf{p}_p, \mathbf{v}_p $ and $\mathbf{a}_p$ ) subtract the current state (including $ \Delta \mathbf{p}_c, \mathbf{v}_c $ and $\mathbf{a}_c$ ).

	The output of the policy network is a $ 3\times 1 $ vector, which is sent to the quadrotor internal attitude and thrust controller directly. 
	
	\section{Path planning of flying throw the gap}
	
	The process of flying through the gap can be split into three stages \cite{falanga2017aggressive}. In the first stage, we compute the traverse trajectory which maximizes the distance between the quadrotor and the edge of the gap. In the second stage, we generate the approach trajectory to guide the drone to fly from the initial hovering position to the desired initial state of the traverse trajectory. In the last stage, we search for a recover trajectory to recover the drone to a hovering state. 
	
	\subsection{Traverse trajectory}
\begin{figure}[h]
		{\includegraphics[width=1.0\columnwidth]{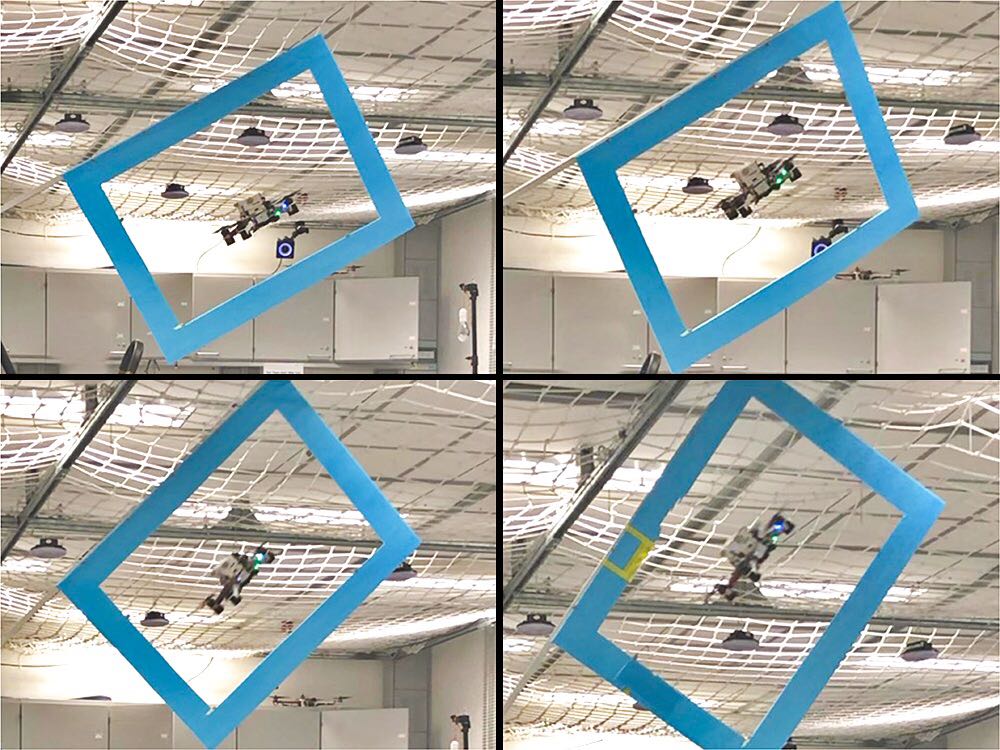}}
		\caption{Our quadrotor flying through narrow gaps with different poses.
			\label{fig:traverse_center}
		}
		\vspace{-0.5cm}
	\end{figure}

	To minimize the risk of collision, we plan our drone flying through the gap's center with its $ Z $-axis orthogonal to the longest side of the gap (Fig. ~\ref{fig:traverse_center}). Our traverse trajectory generation method is the same as \cite{falanga2017aggressive}. 
	
	The traverse trajectory is tracked by a traditional PID controller.
	\subsection{Approach trajectory}
	Once the traverse trajectory is determined, its initial state is the ending state of approach trajectory. Given starting, ending state and traveling time (set as $ 2.6s $ in our work), the optimal motion trajectory can be generated from traditional method \cite{mueller2015computationally} or learning-based method (in Section \ref{Sect_imitate_planning}).
	\subsection{Recover trajectory}
    After crossing the gap, we search a safe recovery trajectory from the drone's current state to a hovering state. The altitude of hover point is set as $ 1 m $ off the ground, its horizontal position is $ 2.5 m$ away from the center of gap in $ X $ direction to leave sufficient clearance. 

	We search the recover trajectory by examining different traveling time ranging from $ 0.5\sim 3.0s $ with a step of $ 0.3 s $ . Once the whole trajectory is within the laboratory size, we exit the searching process and follow the trajectory immediately. Thanks to the computation efficiency of \cite{mueller2015computationally}, we can search a safe trajectory within $ 50ms$.
	
	\section{ Reinforcement learning }
\begin{figure}
		\centering
		\includegraphics[width=0.91\linewidth]{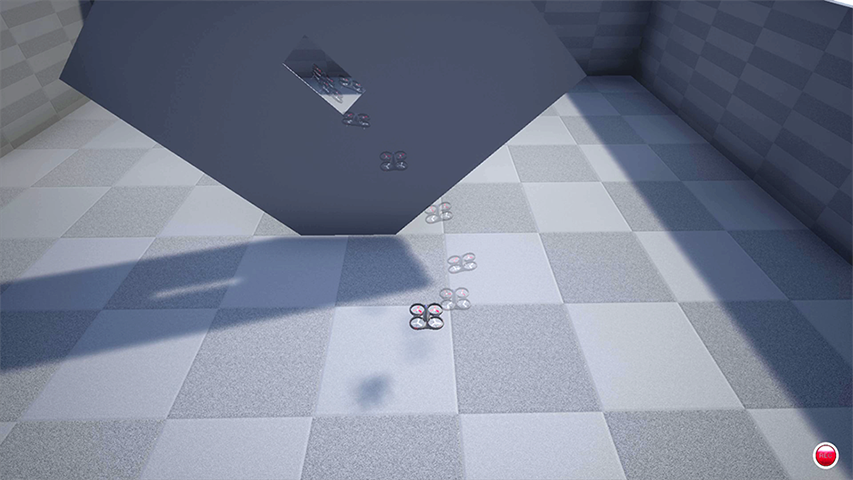}
		\caption{Fine-tuning end-to-end policy network using reinforcement learning in AirSim simulator.}
		\label{fig:airsimmerge}
		\vspace{-1.0cm}
	\end{figure}

	In our work, we fine-tune our end-to-end policy network in \textit{Microsoft}-AirSim simulator \cite{airsim2017fsr} (shown in Fig.~\ref{fig:airsimmerge}). The actual quadrotor parameters are used for the drone model in the AirSim.

	\subsection{Virtual environment setup}\label{Sect:virture_env_setup}
	To improve the generalization ability of the trained network, we train our neural network in different environment settings (different gap poses and drone initial states).
	
	\subsection{Reward function}
		The hand designed reward function in our RL training are divided into two items, the negative (penalty) and positive reward item.
		\subsubsection{Negative reward item}
	    We introduce the penalty term in order to penalize the changes in angular speed, acceleration, and translation acceleration.
			$$
			\begin{aligned}
				R_{neg}(t) =  - &\left(w_{\omega} \norm{\boldsymbol{\omega}(t)} + w_{\alpha} \norm{ \dfrac{d\boldsymbol{\omega}(t)}{dt} } + \right. \\
				 &\left. w_{j}\norm{ \dfrac{d\mathbf{a}(t)}{dt} }  \right) \cdot \Delta t + \mathbf{C} \\
			\end{aligned}
			$$
			where $ w_{j} $,$ w_{\alpha} $ and $ w_{j} $ are the weighting factors, $\boldsymbol{\omega}(t)$ and $\mathbf{a}(t)$  are the angular velocity and linear acceleration, $ \mathbf{C} $ is the collision penalty, $ \Delta t $ is the time interval between current to last sampled time. 
			
			In our work, $ w_{\omega} $,$ w_{\alpha} $ and $ w_{j} $ of penalty item are set to $2\times 57.3$, $ 5 \times 57.3$ and $ 10 $, respectively. If the drone collides with anything (i.e. wall, ground and etc), $ \mathbf{C} $ will be set to $ 10^9 $.
			\subsubsection{Positive reward item}
			If the drone reaches the center of gap, a positive reward will be given
			$$
				R_{pos}(t) = \left( w_{r}\cdot \max(0, d_{a} - \norm{ \mathbf{p}_c - \mathbf{p}(t) } ) \right)\cdot \Delta t + S
			$$
			where $d_{a}$ is the activate distance of positive reward and $ S $ is a one-time reward which occurs at the first time the UAV obtains a positive reward.
			In our work, $d_{a}$ and $w_{r}$ is set to $ 0.15 m $ and $ 1000 $, respectively. $ S $ is set to $ 5\times 10^5 $.
			
	\subsection{RL training}	
		After designing the reward function, we fine-tune our end-to-end policy network using Trust Region Policy Optimization (TRPO) algorithm \cite{schulman2015trust}, which is implemented in \textit{OpenAI}-baselines framework \cite{baselines}.
		
	\section{Results}
	\subsection{Experimental setup}
\begin{figure}[t]
		\subfigure[\label{fig:env_setup} Environmental settup ]
		{\centering\includegraphics[width=0.55\columnwidth]{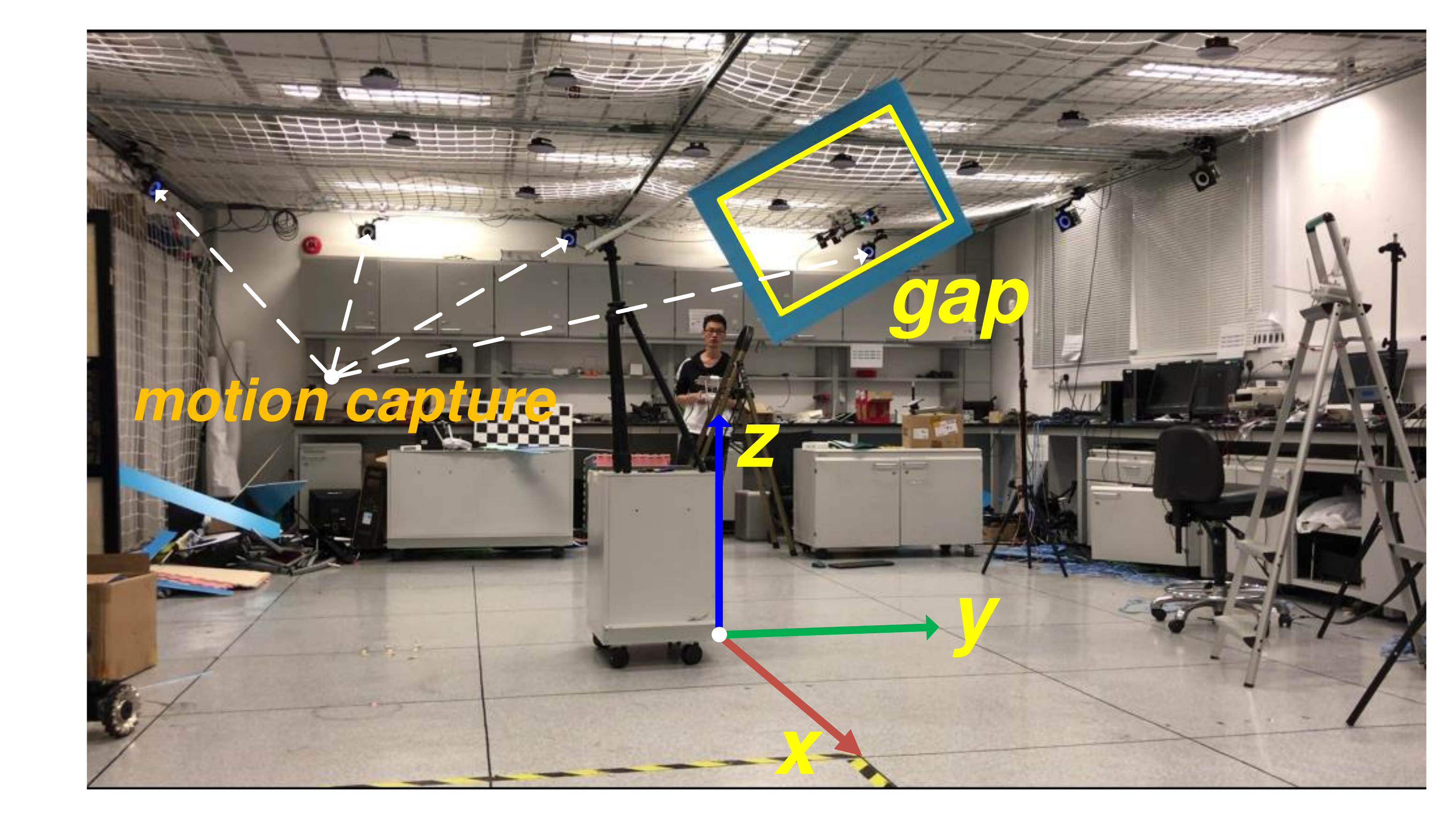}}			
		\subfigure[\label{fig:flying_platfrom} Our flying platfrom ]
		{\centering\includegraphics[width=0.41\columnwidth]{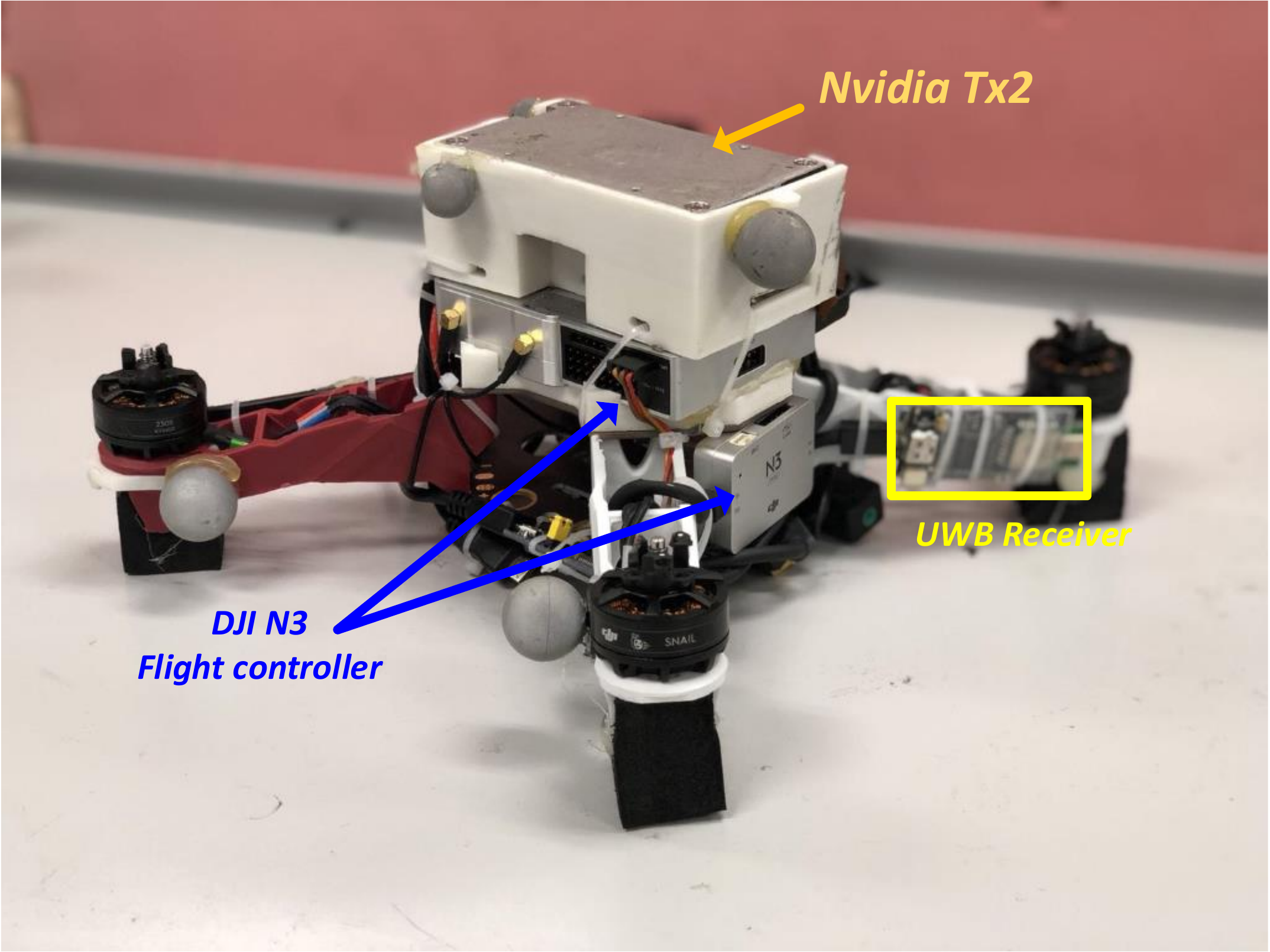}}
		\vspace{-2.0cm}
	\end{figure}

 	The environment settings are shown in Fig.\ref{fig:env_setup}, both the state estimation of quadrotor and gap pose detection is given by motion capture system, which transmits the estimation results to the drone onboard computer via Ultra-WideBand (UWB) wireless module. Our flying platform is show in Fig.~\ref{fig:flying_platfrom}, with \textit{DJI}-N3 as flight controller and \textit{Nvidia}-TX2 as on-board computation platform.
 	The feed-forward process of our planning and end-to-end network cost about $ 3 \sim 5 ms$ and  $ 6\sim 7 ms  $, respectively. 
 	
	\subsection{Imitation of planning}
	\subsubsection{Comparison of different training settings}
	We show the results of different training and learning settings to examine each's effectiveness.
	\begin{itemize}
		\item Setting A: Training without data normalization nor data augmentation.
		\item Setting B: Training without data normalization but with data augmentation.
		\item Setting C: Training with data normalization but without data augmentation.
		\item Setting D: Training with both data normalization augmentation.
	\end{itemize}
	 
	 	\begin{figure}[t]
	 	\centering
	 	{\includegraphics[width=1.0\columnwidth]{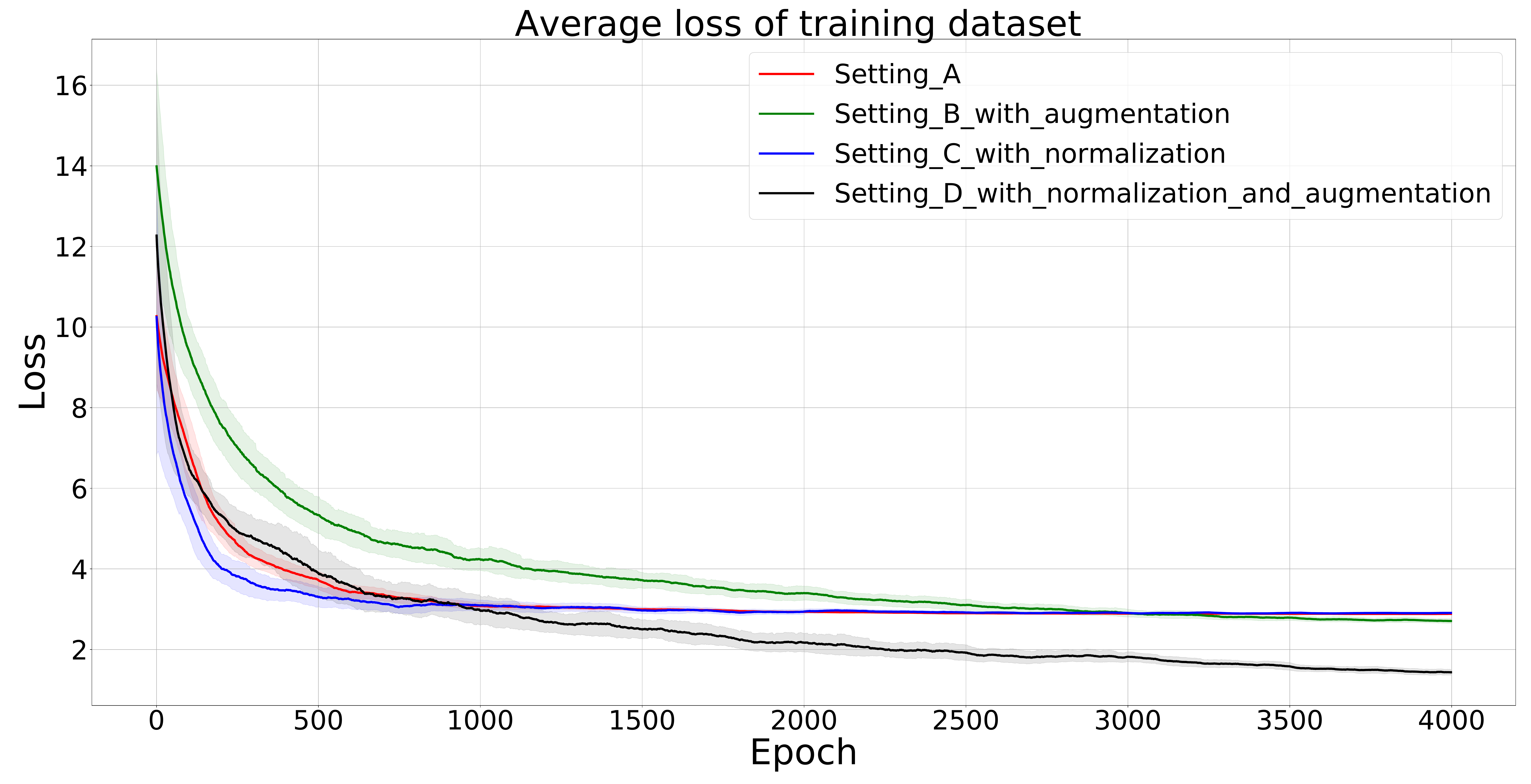}}
	 	\caption{
	 		Comparison of training error:  settings with normalization converge more quickly compare to others and training with data argumentation can achieve a lower loss. The performance of training with both data argumentation and normalization is the best. 
	 		\label{fig:training_loss}}
	 	
	 	{\includegraphics[width=1.0\columnwidth]{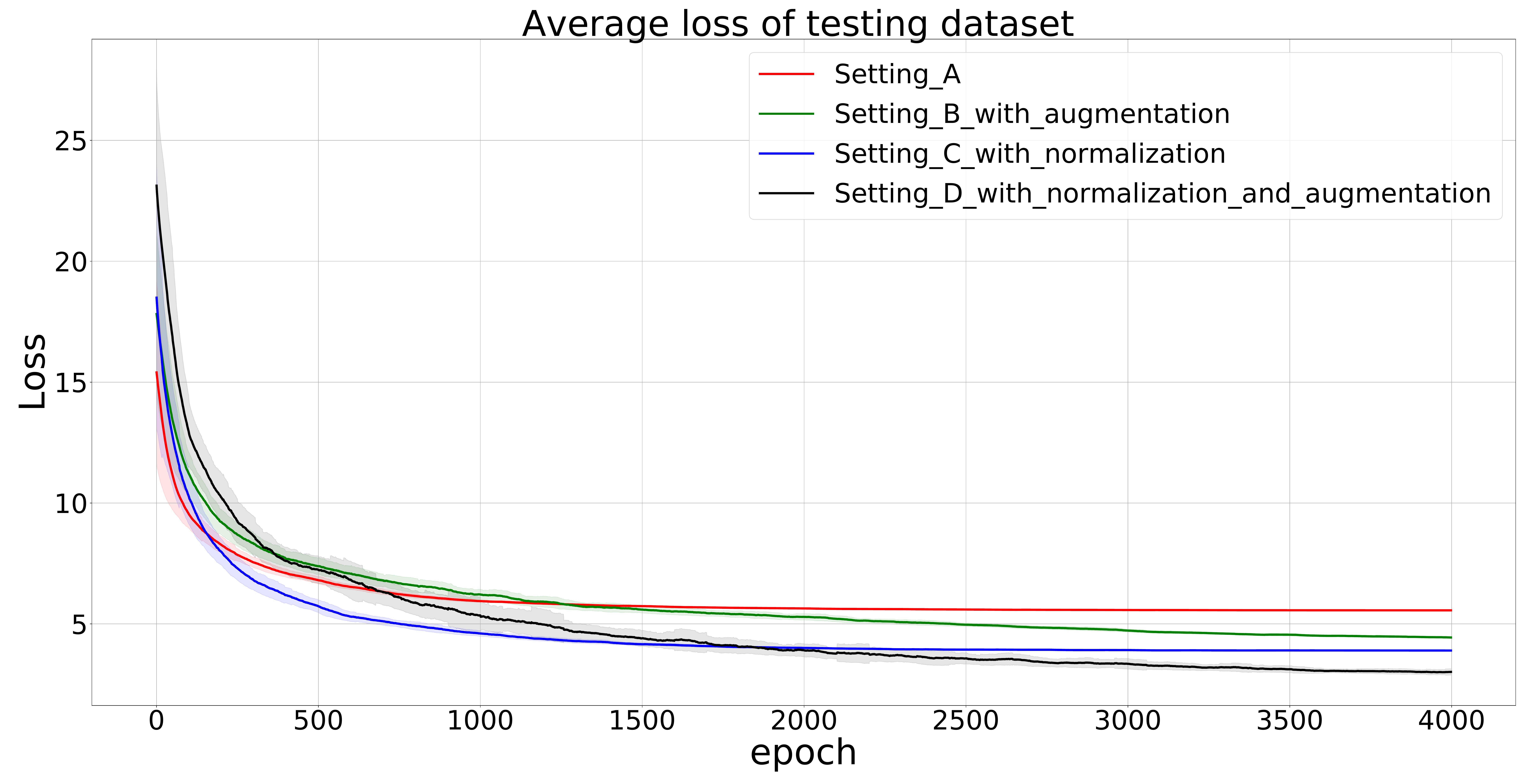}}
	 	\caption{
	 		Comparison of test error: setting with data normalization and argumentation can achieve a lower test loss compare to others, mean that data normalization and argumentation have a positive effect on improving the precision of imitation learning.
	 		\label{fig:test_loss}}
	 	
	 	{\includegraphics[width=1.0\columnwidth]{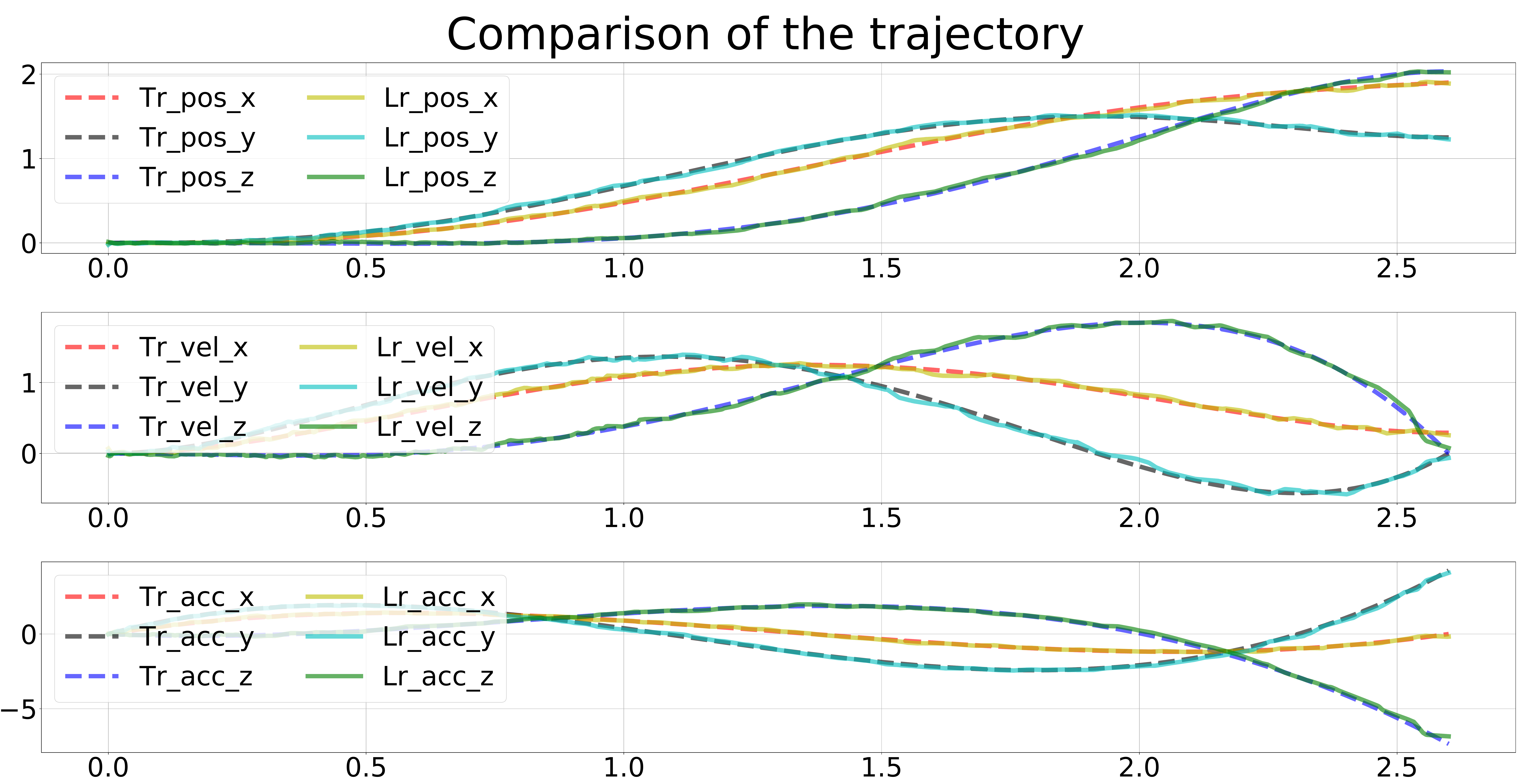}}
	 	\caption{
	 	    Comparison of the trajectory generated from the traditional method (Tr) and learning-based method (Lr). }
	 	\label{fig:planning_network_comp}
	 	\vspace{-0.5cm}
	 \end{figure}
 
 	\begin{figure*}[htbp]
 	\centering
 	\includegraphics[width=1.0\linewidth]{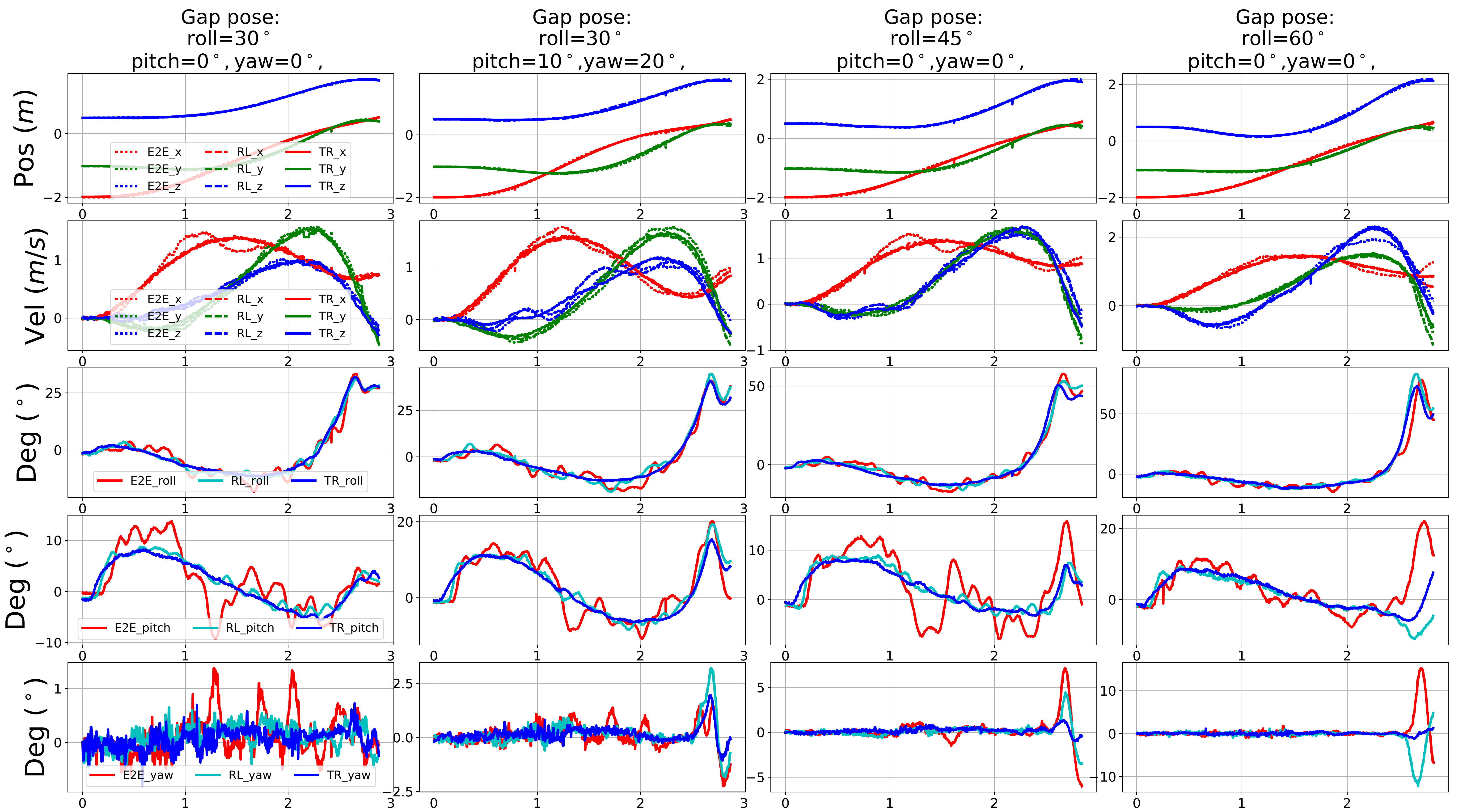}
 	\caption{The trajectory comparison of three types of methods: 1) control using the end-to-end (E2E) policy network. 2) control using reinforcement learning fine-tuned network (RL). 3) control using traditional pipeline (TR)}
 	\label{fig:rlcompare}
 	\subfigure[\label{fig:Average_angular_rate} The comparison of the average angular velocity of three types of methods.]
 	{\includegraphics[width=1.0\columnwidth]{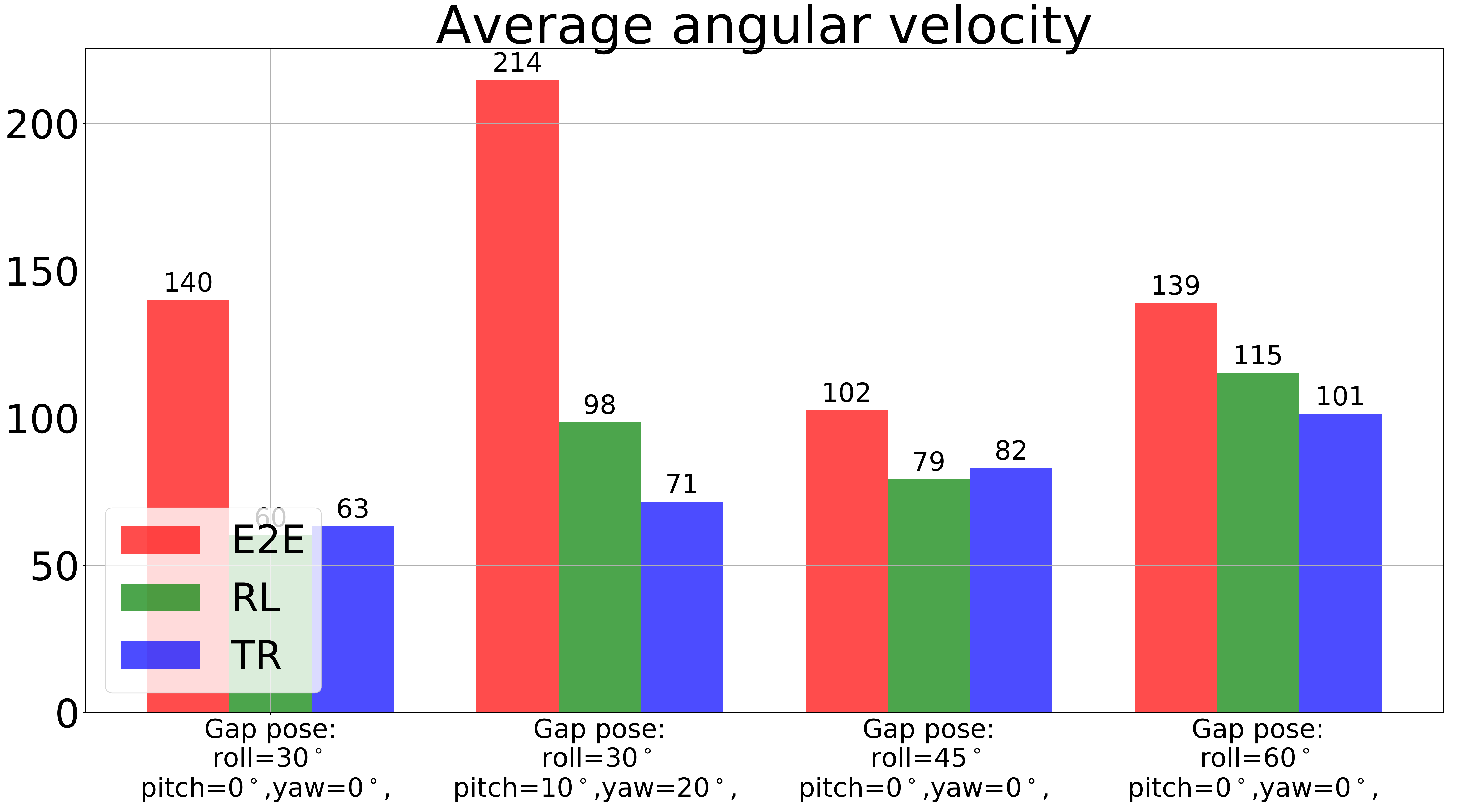}} 
 	\subfigure[\label{fig:Average_thrust} The comparison of the average thrust of three types of methods]
 	{\includegraphics[width=1.0\columnwidth]{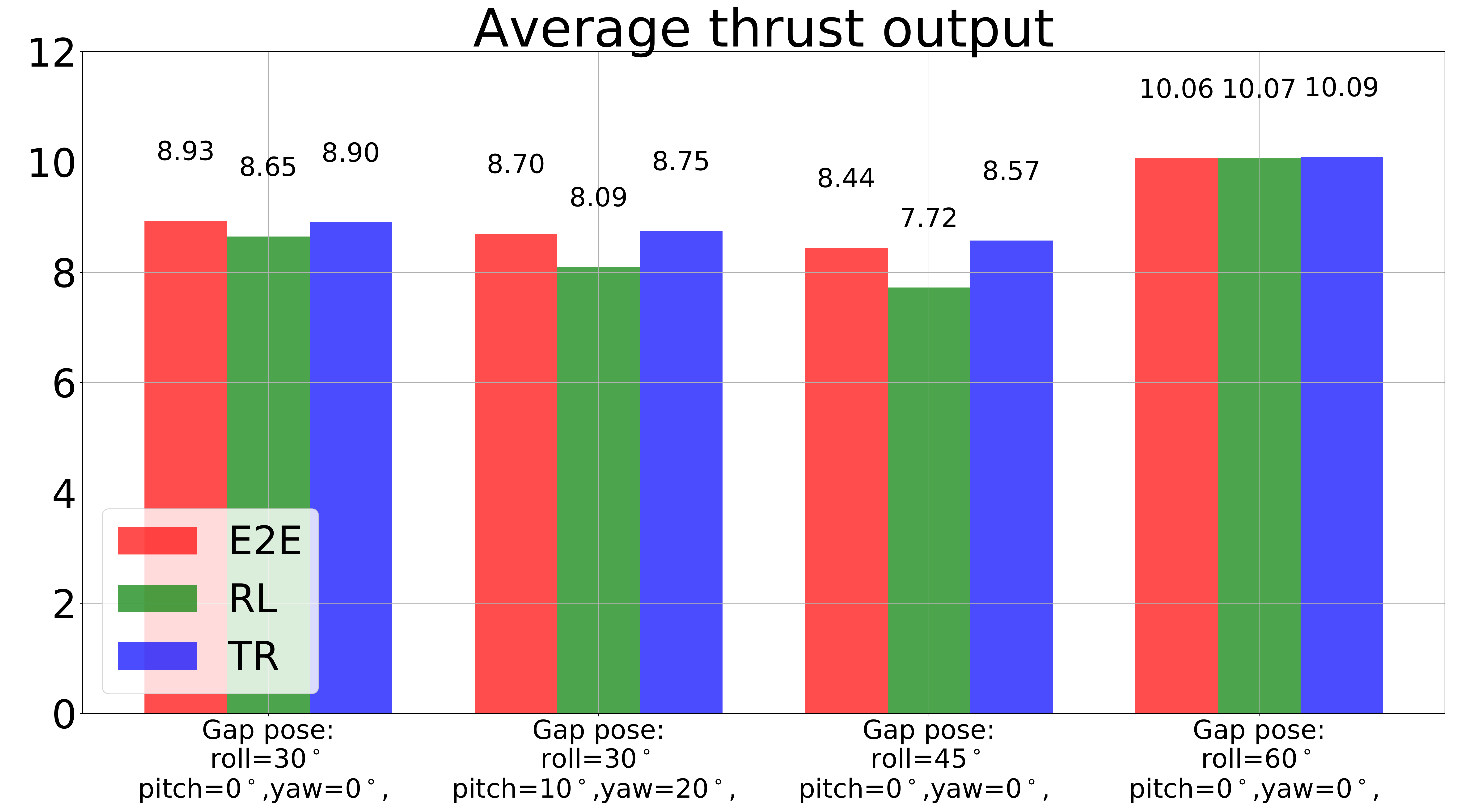}}
 	{\caption{Comparison of average angular velocity and thrust output among three approaches.}}
 	\vspace{-0.5cm}
 \end{figure*}
	
	The four settings have the same learning rate of $ 1.0\times10^{-5} $ and same batch size of 6000, where 4000 data of setting B and D comes from data augmentation(see section.\ref{section_data_argu} ). The training dataset has 20k random trajectories while the testing dataset contains 100 trajectories. Each trajectory contains 1k sample points and the testing dataset is not used for training. After each epoch of training, we compute the average loss of training and testing dataset. The curves of the average loss of training and testing dataset are shown in Fig.~\ref{fig:training_loss} and Fig.~\ref{fig:test_loss}, respectively.
	
	The curve in Fig.~\ref{fig:training_loss} and Fig.~\ref{fig:test_loss} show that both data normalization and argumentation help increase the precision and robustness (generalization) of imitation learning. After training with both data normalization and augmentation in a few days using \textit{Nvidia} GTX 1080ti, our planning network can achieve an average training loss of 0.67 and test loss of 0.99. The comparison of the traditional method and learning method is shown in Fig.~\ref{fig:planning_network_comp}, where we can see the trained neural network imitates the motion primitive well. 
    
	\subsection{Result of imitation learning}\label{sect:res_imitate}
	Although the attitude of the learning based approach are not as smooth as the traditional method, our video and the curve shown in Fig.~\ref{fig:rlcompare} show that the end-to-end method can also successfully cross the gap. 
	
	\subsection{Result of reinforcement learning}
	We fine-tune our neural network using TRPO algorithms \cite{schulman2015trust}, trained in various sets of environment settings Section.~\ref{Sect:virture_env_setup}.  
	
	The comparison of the RL finely turned network, and other learning algorithm are shown in Fig.~\ref{fig:rlcompare}, from which we can see the performance improvement made by RL.
    The comparison of average angular velocity shown in Fig.~\ref{fig:Average_angular_rate} demonstrates that RL algorithm has mitigated the vibration of attitude, which is due to the imitation learning error.
    In Fig.~\ref{fig:Average_thrust}, the average thrust output of RL is the lowest among three types of method, indicating that the neural network is trying the find an efficient way to fly through the gap by consuming lower thrust.  
	
	
	\section{Discussion}
 	In our work, we present an imitate-reinforce training framework, address the problem of flying through a narrow gap using an end-to-end policy network. Our work demonstrates that learning-based approaches can be applied in the area of aggressive-control.   What's more, when compared to the model-based planning and control methods, our neural network fine-tuned by RL consumes lower thrust to accomplish the same mission, indicating that our training framework has the potential to achieve higher performance over the traditional method.
 
 	The future work will be investigating the possibility of a full end-to-end approach in UAV autonomous navigation and control. The current work focuses on the feasibility study of using neural network based control policy, by restricting attention to only planning and control parts. However, the potential of an end-to-end approach lies in improving the perception, mapping, and estimation of model-based methods. 

\newlength{\bibitemsep}\setlength{\bibitemsep}{.0238\baselineskip}
\newlength{\bibparskip}\setlength{\bibparskip}{0pt}
\let\oldthebibliography\thebibliography
\renewcommand\thebibliography[1]{%
	\oldthebibliography{#1}%
	\setlength{\parskip}{\bibitemsep}%
	\setlength{\itemsep}{\bibparskip}%
}

\bibliography{rss2019jiarong}

\begin{thebibliography}{10}
\providecommand{\url}[1]{#1}
\csname url@rmstyle\endcsname
\providecommand{\newblock}{\relax}
\providecommand{\bibinfo}[2]{#2}
\providecommand\BIBentrySTDinterwordspacing{\spaceskip=0pt\relax}
\providecommand\BIBentryALTinterwordstretchfactor{4}
\providecommand\BIBentryALTinterwordspacing{\spaceskip=\fontdimen2\font plus
\BIBentryALTinterwordstretchfactor\fontdimen3\font minus
  \fontdimen4\font\relax}
\providecommand\BIBforeignlanguage[2]{{%
\expandafter\ifx\csname l@#1\endcsname\relax
\typeout{** WARNING: IEEEtran.bst: No hyphenation pattern has been}%
\typeout{** loaded for the language `#1'. Using the pattern for}%
\typeout{** the default language instead.}%
\else
\language=\csname l@#1\endcsname
\fi
#2}}

\bibitem{falanga2017aggressive}
D.~Falanga, E.~Mueggler, M.~Faessler, and D.~Scaramuzza, ``Aggressive quadrotor
  flight through narrow gaps with onboard sensing and computing using active
  vision,'' in \emph{Robotics and Automation (ICRA), 2017 IEEE International
  Conference on}.\hskip 1em plus 0.5em minus 0.4em\relax IEEE, 2017, pp.
  5774--5781.

\bibitem{loianno2017estimation}
G.~Loianno, C.~Brunner, G.~McGrath, and V.~Kumar, ``Estimation, control, and
  planning for aggressive flight with a small quadrotor with a single camera
  and imu,'' \emph{IEEE Robotics and Automation Letters}, vol.~2, no.~2, pp.
  404--411, 2017.

\bibitem{mueller2015computationally}
M.~W. Mueller, M.~Hehn, and R.~D'Andrea, ``A computationally efficient motion
  primitive for quadrocopter trajectory generation,'' \emph{IEEE Transactions
  on Robotics}, vol.~31, no.~6, pp. 1294--1310, 2015.

\bibitem{lin2018autonomous}
Y.~Lin, F.~Gao, T.~Qin, W.~Gao, T.~Liu, W.~Wu, Z.~Yang, and S.~Shen,
  ``Autonomous aerial navigation using monocular visual-inertial fusion,''
  \emph{Journal of Field Robotics}, vol.~35, no.~1, pp. 23--51, 2018.

\bibitem{gaoflying}
F.~Gao, W.~Wu, W.~Gao, and S.~Shen, ``Flying on point clouds: Online trajectory
  generation and autonomous navigation for quadrotors in cluttered
  environments,'' \emph{Journal of Field Robotics}.

\bibitem{hwangbo2019learning}
J.~Hwangbo, J.~Lee, A.~Dosovitskiy, D.~Bellicoso, V.~Tsounis, V.~Koltun, and
  M.~Hutter, ``Learning agile and dynamic motor skills for legged robots,''
  \emph{Science Robotics}, vol.~4, no.~26, p. eaau5872, 2019.

\bibitem{giusti2016machine}
A.~Giusti, J.~Guzzi, D.~C. Ciresan, F.-L. He, J.~P. Rodr{\'\i}guez, F.~Fontana,
  M.~Faessler, C.~Forster, J.~Schmidhuber, G.~Di~Caro, \emph{et~al.}, ``A
  machine learning approach to visual perception of forest trails for mobile
  robots.'' \emph{IEEE Robotics and Automation Letters}, vol.~1, no.~2, pp.
  661--667, 2016.

\bibitem{jung2018perception}
S.~Jung, S.~Hwang, H.~Shin, and D.~H. Shim, ``Perception, guidance, and
  navigation for indoor autonomous drone racing using deep learning,''
  \emph{IEEE Robotics and Automation Letters}, vol.~3, no.~3, pp. 2539--2544,
  2018.

\bibitem{kaufmann2018deep}
E.~Kaufmann, A.~Loquercio, R.~Ranftl, A.~Dosovitskiy, V.~Koltun, and
  D.~Scaramuzza, ``Deep drone racing: Learning agile flight in dynamic
  environments,'' \emph{arXiv preprint arXiv:1806.08548}, 2018.

\bibitem{kaufmann2018beauty}
E.~Kaufmann, M.~Gehrig, P.~Foehn, R.~Ranftl, A.~Dosovitskiy, V.~Koltun, and
  D.~Scaramuzza, ``Beauty and the beast: Optimal methods meet learning for
  drone racing,'' \emph{arXiv preprint arXiv:1810.06224}, 2018.

\bibitem{abbeel2007application}
P.~Abbeel, A.~Coates, M.~Quigley, and A.~Y. Ng, ``An application of
  reinforcement learning to aerobatic helicopter flight,'' in \emph{Advances in
  neural information processing systems}, 2007, pp. 1--8.

\bibitem{mellinger2012trajectory}
D.~Mellinger, N.~Michael, and V.~Kumar, ``Trajectory generation and control for
  precise aggressive maneuvers with quadrotors,'' \emph{The International
  Journal of Robotics Research}, vol.~31, no.~5, pp. 664--674, 2012.

\bibitem{hornik1989multilayer}
K.~Hornik, M.~Stinchcombe, and H.~White, ``Multilayer feedforward networks are
  universal approximators,'' \emph{Neural networks}, vol.~2, no.~5, pp.
  359--366, 1989.

\bibitem{cybenko1989approximation}
G.~Cybenko, ``Approximation by superpositions of a sigmoidal function,''
  \emph{Mathematics of control, signals and systems}, vol.~2, no.~4, pp.
  303--314, 1989.

\bibitem{goodfellow2016deep}
I.~Goodfellow, Y.~Bengio, A.~Courville, and Y.~Bengio, \emph{Deep
  learning}.\hskip 1em plus 0.5em minus 0.4em\relax MIT press Cambridge, 2016,
  vol.~1.

\bibitem{sola1997importance}
J.~Sola and J.~Sevilla, ``Importance of input data normalization for the
  application of neural networks to complex industrial problems,'' \emph{IEEE
  Transactions on Nuclear Science}, vol.~44, no.~3, pp. 1464--1468, 1997.

\bibitem{ioffe2015batch}
S.~Ioffe and C.~Szegedy, ``Batch normalization: Accelerating deep network
  training by reducing internal covariate shift,'' \emph{arXiv preprint
  arXiv:1502.03167}, 2015.

\bibitem{singh2015investigations}
B.~K. Singh, K.~Verma, and A.~Thoke, ``Investigations on impact of feature
  normalization techniques on classifier's performance in breast tumor
  classification,'' \emph{International Journal of Computer Applications}, vol.
  116, no.~19, 2015.

\bibitem{mellinger2011minimum}
D.~Mellinger and V.~Kumar, ``Minimum snap trajectory generation and control for
  quadrotors,'' in \emph{Robotics and Automation (ICRA), 2011 IEEE
  International Conference on}.\hskip 1em plus 0.5em minus 0.4em\relax IEEE,
  2011, pp. 2520--2525.

\bibitem{lee2010geometric}
T.~Lee, M.~Leok, and N.~H. McClamroch, ``Geometric tracking control of a
  quadrotor uav on se (3),'' in \emph{49th IEEE conference on decision and
  control (CDC)}.\hskip 1em plus 0.5em minus 0.4em\relax IEEE, 2010, pp.
  5420--5425.

\bibitem{airsim2017fsr}
\BIBentryALTinterwordspacing
S.~Shah, D.~Dey, C.~Lovett, and A.~Kapoor, ``Airsim: High-fidelity visual and
  physical simulation for autonomous vehicles,'' in \emph{Field and Service
  Robotics}, 2017. [Online]. Available: \url{https://arxiv.org/abs/1705.05065}
\BIBentrySTDinterwordspacing

\bibitem{schulman2015trust}
J.~Schulman, S.~Levine, P.~Abbeel, M.~I. Jordan, and P.~Moritz, ``Trust region
  policy optimization.'' in \emph{Icml}, vol.~37, 2015, pp. 1889--1897.

\bibitem{baselines}
P.~Dhariwal, C.~Hesse, O.~Klimov, A.~Nichol, M.~Plappert, A.~Radford,
  J.~Schulman, S.~Sidor, Y.~Wu, and P.~Zhokhov, ``Openai baselines,''
  \url{https://github.com/openai/baselines}, 2017.

\end{thebibliography}
	
\end{document}